\documentclass[letterpaper, 10 pt, conference]{ieeeconf}  

\IEEEoverridecommandlockouts                              

\usepackage{amsmath} 

\usepackage{diagbox}
\usepackage{graphicx}
\usepackage{wrapfig}
\usepackage{multirow} 
\usepackage{stfloats}
\usepackage{xcolor}
\usepackage{tabularx}

\usepackage{amsmath}
\usepackage{amssymb}

\usepackage{pgfplots}
\pgfplotsset{compat=1.18}

\newtheorem{assumption}{Assumption}

\title{\LARGE \bf
Good in Bad (GiB): Sifting Through End-user Demonstrations for Learning a Better Policy
}

\author{Noushad Sojib$^{1}$, Ola Ghattas$^{1}$, Sajay Arthanat$^{2}$ and Momotaz Begum$^{1}$
\thanks{$^{1}$Noushad Sojib, Ola Ghattas and Momotaz Begum are with the Department of Computer Science, University of New Hampshire, USA {\tt\small noushad.sojib@unh.edu},{\tt\small ola.ghattas@unh.edu}, {\tt\small mbegum@cs.unh.edu}}%
\thanks{$^{2}$Sajay Arthanat is with the Department of Occupational Therapy and the Center for Digital Health Innovation, University of New Hampshire, USA {\tt\small sajay.arthanat@unh.edu}}%
}

\begin{document}

\maketitle
\thispagestyle{empty}
\pagestyle{empty}

\begin{abstract}

Imitation learning offers a promising framework for enabling robots to acquire diverse skills from human users. However, most imitation learning algorithms assume access to high-quality demonstrations—an unrealistic expectation when collecting data from non-expert users, whose demonstrations often contain inadvertent errors. Naively learning from such demonstrations can result in unsafe policy behavior, while discarding entire demonstrations due to occasional mistakes wastes valuable data, especially in low-data settings. In this work, we introduce \textbf{GiB (Good-in-Bad)}, an algorithm that automatically identifies and discards erroneous subtasks within demonstrations while preserving high-quality subtasks. The filtered data can then be used by any policy learning algorithm to train more robust policies. GiB first trains a self-supervised model to learn latent features and assigns binary weights to label each demonstration as good or bad. It then models the latent feature distribution of high-quality segments and uses the Mahalanobis distance to detect and evaluate poor-quality subtasks. We validate GiB on the Franka robot in both simulated and real-world multi-step tasks, demonstrating improved policy performance when learning from mixed-quality human demonstrations.
\end{abstract}

\section{INTRODUCTION}


Imitation Learning (IL) has seen rapid progress in controlled laboratory settings, with numerous policy learning approaches showcased at top robotics and machine learning conferences~\cite{zare2024survey} trained on top of high-quality demonstrations. However, to translate these advances into real-world impact, especially to empower lay users to teach robots new tasks, there is a growing need to extend IL into field settings, a direction we referred to as Field Imitation Learning (FIL). FIL introduces unique challenges, particularly those stemming from the variability and imperfection of human-provided demonstrations. A key open problem is how to handle \textit{human errors}, inadvertent mistakes made by non-expert users due to unfamiliarity with robots or fatigue during long demonstration sessions. This challenge has only recently begun to gain attention in the IL literature~\cite{belkhale2024data}. This paper investigates a central question: \textit{How can we best utilize a limited set of demonstrations from non-expert users when some are corrupted by human error?}

Policy learning performance is tightly coupled with the quantity and quality of training data~\cite{belkhale2024data}. Yet, much of contemporary IL research assumes access to abundant, high-quality demonstrations either generated in simulation~\cite{hua2021learning} or by expert users~\cite{ross2011reduction,hussein2017imitation}. These assumptions rarely hold in practical settings, particularly for complex, multi-step household tasks like making coffee, cleaning up, or operating appliances. In such cases, a single mistake in one subtask can undermine the validity of the entire demonstration, placing significant physical and cognitive demands on end-users.

As human demonstrations are expensive, discarding entire demonstrations due to partial errors is often infeasible. Addressing this, we propose \textit{GiB} (Good-in-Bad), an algorithmic framework designed to identify and retain high-quality segments from imperfect human demonstrations while filtering out those that may induce unsafe policies.

GiB operates in two stages. The first stage distinguishes good and bad demonstrations by penalizing inconsistencies, similar to Behavior Cloning for Error Discovery (BED) \cite{sojib2024self}. The second stage goes further by analyzing erroneous demonstrations at the subtask level, leveraging Mahalanobis distance to isolate useful segments. GiB is designed specifically for sequential visual tasks, and Figure~\ref{fig:gib} provides an overview of its architecture.

The main contributions of this work are as follows: 1) We introduce \textit{GiB}, a framework designed to tackle the overlooked challenge of learning from non-expert demonstrations corrupted by human errors. By analyzing visual demonstrations of multi-step tasks, GiB automatically detects and filters unreliable segments, yielding high-quality training data suitable for any imitation learning algorithm. 2) We perform extensive evaluations in both simulation and on a real Franka robot with demonstrations collected by lay users. Our results show that GiB-curated datasets consistently yield significantly higher policy performance compared to unfiltered or alternatively filtered baselines.


\section{Related Work}
Recent studies have discussed the role of diversity in improving the generalizability of IL policies over the quantity of data \cite{robomimic2021, mandlekar2023mimicgen}. For instance, having a diverse set of demonstrations is more beneficial for policy learning than simply increasing the number of demonstrations \cite{belkhale2024data}. This insight has brought the research on leveraging sub-optimal human demonstrations for policy learning into the limelight. We focus on offline imitation learning and discuss existing efforts from three perspectives: a) data filtering, b) policy learning from suboptimal demonstrations, and c) anomaly detection methods.

\textbf{Filtering out bad demonstrations: }While diverse datasets support robust policy learning, noisy or suboptimal demonstrations can hinder performance. RoboMimic \cite{robomimic2021} shows that excluding the worst-quality demonstrations from a multi-human, mixed-quality dataset leads to improved policy performance. S2I \cite{chen2024towards} uses preference learning to rank subtasks but requires a fixed set of good demonstrations a priori, limiting practicality. DemInf \cite{hejna2025robot} estimates demonstration quality using mutual information; however, it cannot assess the quality of individual subtasks.  Offline-ILID \cite{yue2024leverage} selects data based on resultant states to capture expert and diverse behaviors more effectively, yet still relies on labeled demonstrations. In contrast, GiB assumes good demonstrations are more prevalent and autonomously identifies high-quality segments without any external labels or curated expert sets. Imitation Learning from Purified Demonstrations \cite{wang2023imitation} refines poor segments using diffusion models, but these are impractical for vision-based tasks due to high data demands. Robust IL \cite{hussein2021robust} down-weights noisy data during training but is limited to discrete action spaces. The BED framework \cite{sojib2024self}, used in GiB, discards entire trajectories if any error is found, preserving quality but reducing data efficiency. GiB instead refines data at the segment level, preserving usable information and improving learning.

\textbf{Leveraging sub-optimal demonstrations for policy learning: }Few IL approaches directly address learning from suboptimal demonstrations \cite{li2024imitation, kim2022demodice}. BCND \cite{sasaki2020behavioral}, an extension of behavior cloning, implicitly downweights noisy actions by favoring frequent behaviors, offering robustness to occasional errors without explicit labeling. However, this frequency bias limits generalization in complex tasks and struggles with convergence in high-dimensional, vision-based settings. Other recent methods \cite{hoque2024intervengen, gandhi2023eliciting} improve robustness by diversifying demonstrations through active interventions during data collection. In contrast, GiB is policy-agnostic and focuses on curating data for downstream use with any state-of-the-art policy.

\textbf{Anomaly detection methods: }Identifying suboptimal segments in demonstrations is analogous to outlier detection in machine learning, with methods such as KNN \cite{ramaswamy2000efficient}, DeepSVDD \cite{ruff2018deep}, IForest \cite{liu2008isolation}, and LOF \cite{breunig2000lof} being state-of-the-art. However, applying these to imitation learning is non-trivial due to two key challenges: (1) IL data is sequential, violating the i.i.d. assumption of most anomaly detectors; and (2) the high dimensionality and variable-length segments typical in IL tasks are incompatible with algorithms designed for low-dimensional, fixed-size inputs. While OIL-AD \cite{wang2024oil} takes a step toward integrating anomaly detection with IL, its effectiveness remains limited in complex and realistic tasks. On the contrary GiB is designed to detect anomalies in sequential data where each data point can vary in length.


\section{Preliminaries}

\textbf{Preliminaries on Imitation Learning: }
We consider imitation learning as a Markov Decision Process (MDP) represented by the tuple \((S, A, r, q, s_0, \gamma)\). Here, the reward function \(r\) and the environment dynamics \(q\) are unknown. However, we assume access to a dataset \(D = \{\tau_1, \tau_2, \ldots, \tau_N\}\) comprising of \(N\) trajectories, which include both optimal and erroneous demonstrations. Every trajectory \(\tau_i\) is a sequence of state-action pairs, \(\tau_i = \{(s_1, a_1), (s_2, a_2), \ldots, (s_{T_i}, a_{T_i})\}\), where \(s \in S\) denotes a state, \(a \in A\) denotes an action, and \(T_i\) is the horizon of the trajectory \(\tau_i\). In case of behavior cloning (BC), the objective is to learn a policy \(\pi_\theta : S \rightarrow A\) that maps states to actions by minimizing the standard BC loss function $\mathbb{E}_{(s,a)\sim D}[-\log \pi_{\theta}(a|s)] $.

\textbf{Weighted policy learning:}
When learning from imperfect demonstrations, a common strategy is to assign weights to individual (state, action) pairs within the dataset $D$ \cite{xu2022discriminator}. Alternatively, weights can be assigned at the level of entire trajectories or their segments. For example, in our prior work [reference hidden], we modified the standard BC loss function for weighted policy learning as follows:

\begin{equation}
\mathcal{L} = - \frac{1}{N} \sum_{i=1}^{N} w_i \sum_{(s,a)\sim \tau_i} \log \pi_\theta (a|s)
\label{eqn: prob1}
\end{equation}

The weights $w_i$ are learned directly from demonstrations using a neural network. The proposed GiB framework builds on this idea by assigning weights to individual subtasks within a demonstration according to their correctness.


\section{Good in Bad (GiB): An Algorithm for Detecting Good Segments in Corrupted Demonstrations} 
Given a dataset \( D \) composed of mixed-quality demonstrations, we make the following assumption:

\begin{assumption}
Good demonstrations outnumber suboptimal ones. 

\end{assumption}

This assumption ensures learning feasibility in label-free settings: without explicit quality labels, the algorithm implicitly identifies dominant patterns. If poor demonstrations dominate, the model is more likely to learn undesirable behavior.

\textbf{Problem Formulation: } Assuming each task trajectory can be partitioned into a maximum number of segments $k$, we define the loss function for subtask-weighted policy learning as follows:

\begin{multline}
\mathcal{L} = - \frac{1}{N} \sum_{i=1}^{N} \sum_{j=1}^k \beta_{ij}
    \sum_{(s,a) \sim \tau_i^j} \log \pi_\theta(a \mid s) \\
\text{subject to} \quad
\sum_{i=1}^{N} \sum_{j=1}^{k} \beta_{ij} = N \cdot k - \rho
\label{eqn: prob2}
\end{multline}


Here, $\beta_{ij} \in \{0,1\}$  denotes the binary weight for the $j$-th subtask of the $i$-th trajectory, $\tau_i^j$. The policy $\pi_\theta$ can be any SOTA algorithm. The hyperparameter $\rho$ specifies the number of subtasks to be removed from the entire dataset. Unlike Equation~\ref{eqn: prob1}, which assigns a weight $w_i$ to an entire trajectory, Equation~\ref{eqn: prob2} operates at the subtask level, assigning weights to individual segments. GiB aims to learn these binary weights $\beta_{ij}$ to retain high-quality segments and discard low-quality ones. We use the terms \textit{subtask} and \textit{segment} interchangeably throughout the paper.

\begin{figure*}[thpb]
  \centering
  \includegraphics[scale=0.5]{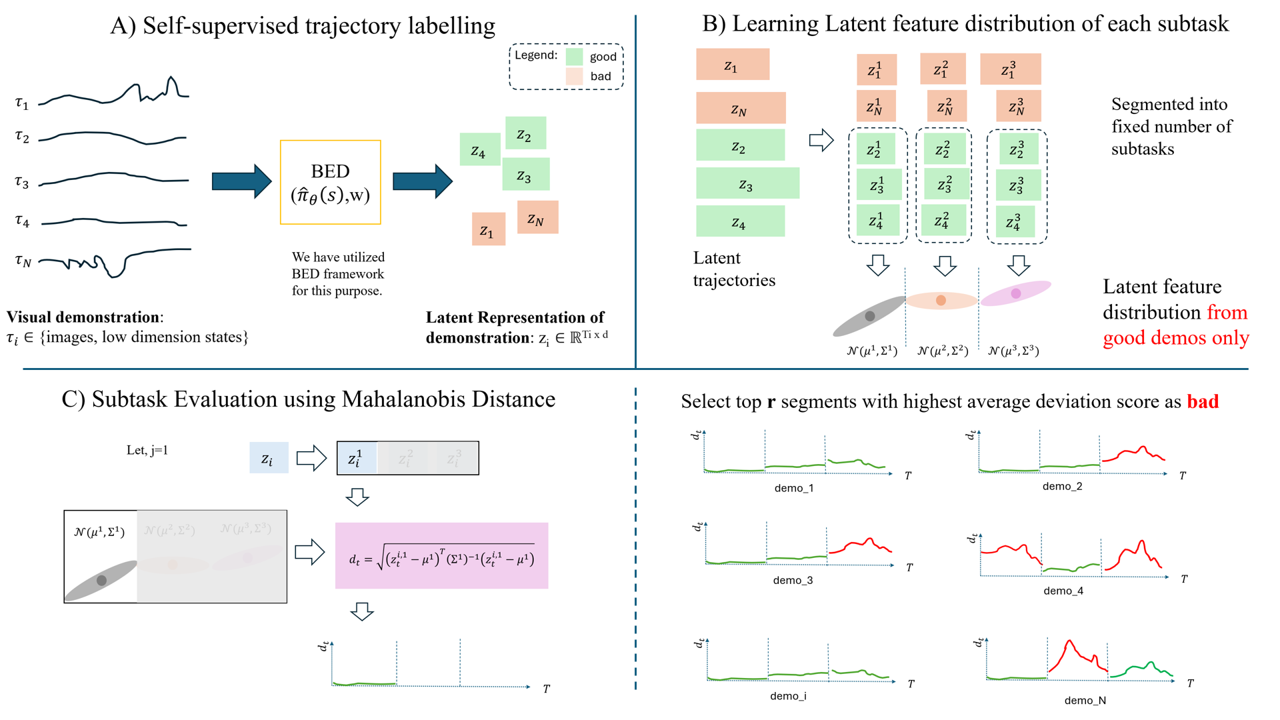}
  \caption{GiB subtask evaluation pipeline.
 (A) A self-supervised model learns latent embeddings \( z_i \) and assigns binary weights \( w_i \in \{0, 1\} \) to trajectories, with \( w_i = 1 \) indicating high-quality demonstrations. (B) Trajectories are segmented into \( k \) subtasks (e.g. $k=3$), and for each subtask \( j \), a Gaussian distribution of the latent trajectory \( \mathcal{N}(\mu^j, \Sigma^j) \) is estimated using high-quality trajectories only.  (C) Subtasks are evaluated by computing the Mahalanobis distance between each timestep’s latent embedding and the corresponding subtask distribution. The top 
$\rho$ subtasks with the largest deviation scores are labeled as bad (shown in red), while the remaining ones are labeled as good (shown in green).
}
  \label{fig:gib}
\end{figure*}

GiB operates in two stages: self-supervised representation learning followed by subtask evaluation. In the first stage, GiB learns latent feature representations by minimizing an inconsistency loss, as defined in Equation~\ref{eqn: obj_path}. This stage assigns a trajectory weight $w_i$ to each demonstration, where $w_i = 1$ indicates a good-quality trajectory and $w_i = 0$ denotes a poor-quality one. In the second stage, GiB models the latent feature distribution for each subtask using only the good trajectories ($w_i = 1$) identified in the first stage.  Then, for all subtasks across all trajectories, each subtask is evaluated by computing its Mahalanobis distance from the corresponding feature distribution. A deviation score is assigned to each subtask based on this distance: larger distances indicate greater deviation from the expected good behavior. Subtasks are ranked according to their average deviation scores. The top $\rho$ most deviating subtasks (i.e., those with the highest distances) are assigned $\beta_{ij} = 0$ (bad), while the remaining subtasks are assigned $\beta_{ij} = 1 (good)$. Figure \ref{fig:gib} summarizes the concept.

\subsection{Self-supervised feature learning}

A demonstration is considered bad if it is inconsistent with the majority of the demonstrations (see Assumption 1). We consider three types of inconsistencies: a) Action inconsistency: the action at a given state differs from the majority of similar states. b) State inconsistency: the state evolution deviates from the dominant trajectory patterns. c) Goal inconsistency: the demonstration ends at a state different from the common end states.
We adopt a self-supervised approach that learns the latent feature distribution of states and simultaneously learns to weight each demonstration as good or bad. Specifically, we build upon the BED framework \cite{sojib2024self}, which defines a loss function based on these inconsistencies:

\begin{equation}
\begin{split}
    \mathcal{L}(D,w,m) = &  c \cdot \frac{1}{N} \cdot
    \sum_{i=1}^N w_i   \cdot  \sum_{(s,a) \sim \tau_i} \big ( \hat{\pi}_\theta (s) - a \big  ) ^2 \\
    & + h \cdot \sum_{i=1}^N w_i \cdot \|\mathbf{G}- \mathbf{g_i} \| 
    \\
   & + q \cdot \sum_{i=1}^N w_i \cdot \|\mathbf{Z}- \mathbf{z_i}\| 
    + k \cdot \left(m\cdot N-\sum_{j=1}^{N} w_j \right)^2 \\
\end{split}
\label{eqn: obj_path}
\end{equation}



Here, $\hat{\pi}_{\theta}$ is a self-supervised BC policy that employs a feature encoder $\phi$ to map states to $d$-dimensional latent representations and predict corresponding actions $\hat{\pi}_{\theta}(s)$. The hyperparameters $c$, $h$, and $q$ weigh the contributions of action consistency, goal consistency, and path consistency terms, respectively. The variable $m$ denotes the expected number of high-quality demonstrations, while $k$ controls the strength of the soft constraint enforcing this prior.

Although the nominal goal $\mathbf{G} \in \mathbb{R}^d$ is not directly observed, we approximate it using the weighted average of the terminal latent states $\mathbf{g}_i \in \mathbb{R}^d$ of each demonstration. Likewise, the nominal latent path $\mathbf{Z} \in \mathbb{R}^{T \times d}$ is estimated as the expected latent trajectory over these demonstrations, where each $\mathbf{z}_i \in \mathbb{R}^{|\tau_i| \times d}$ represents the encoded trajectory corresponding to demonstration $i$. (See Appendix for the exact formulations.)

Demonstrations with higher inconsistencies naturally incur larger losses. As a result, to minimize the overall objective during training, the optimizer assigns lower weights (i.e., $w \approx 0$) to inconsistent trajectories and higher weights (i.e., $w \approx 1$) to consistent ones. In practice, the learned weights are clipped to the $[0, 1]$ range and interpreted as approximate binary assignments.

\subsection{Identifying Good Subtasks in Demonstrations}
We split each trajectory into smaller subtasks and then label each subtask as “good” or “bad” based on how consistent it is in the latent space.

\textbf{Subtask segmentation: } Humans typically perceive long-horizon tasks as a sequence of discrete object interactions. For instance, the task of making coffee can be naturally decomposed into subtasks such as picking up a coffee pod, placing it into the machine. Motivated by this observation, we adopt an object-centric segmentation strategy for subtask generation. In simulation, we leverage the MimicGen framework, which uses task-specific configuration files to automatically generate subtask annotations. For real-world robot experiments, we used a heuristic-based script to segment demonstrations into subtasks directly from low-level trajectory data (end-effector poses and gripper states). The script detects cues such as gripper opening or the end-effector exceeding a height threshold, removing the need for manual labeling. We assume each trajectory can be divided into $k$ number of subtasks. We use the subtask annotation to divide each latent trajectory $z_i \in  \mathbb{R}^{T_i\times d}$ into corresponding latent subtask $z_i^{j\in k} \in \mathbb{R}^{T_i^j \times d}$.

\textbf{Subtask Evaluation:} Task execution exhibits high variability, especially in high-dimensional state spaces. This variation is even more pronounced in end-user demonstrations, which may include both valid strategic differences and suboptimal behaviors. To assess subtask quality effectively, a consistency metric must distinguish meaningful variations from noise. Rather than comparing trajectories to a fixed reference, we measure deviation from a learned distribution of valid behaviors. For each subtask \( j \), we model the distribution of state features across all demonstrations using a multivariate Gaussian \( \mathcal{N}(\mu^j, \Sigma^j) \), where \( \mu^j \in \mathbb{R}^d \) is the mean and \( \Sigma^j \in \mathbb{R}^{d \times d} \) is the covariance matrix. These statistics are computed from the set of all feature vectors in subtask \(j\) across \(N\) trajectories with weight $w_i=1$, each with variable length \(T^j_i\):

\begin{equation}
\mathcal{X}^j = \left\{ z_t^{i,j} \;\middle|\; i = 1, \dots, N,\; t = 1, \dots, T_i^j,\; w[i] = 1 \right\}
\end{equation}

\begin{equation}
\mu^j = \frac{1}{|\mathcal{X}^j|} \sum_{x \in \mathcal{X}^j} x, \quad
\Sigma^j = \frac{1}{|\mathcal{X}^j| - 1} \sum_{x \in \mathcal{X}^j} (x - \mu^j)(x - \mu^j)^\top
\end{equation}

Note that the distribution \( \mathcal{N}(\mu^j, \Sigma^j) \) is defined in the latent space and is invariant to the temporal duration of each subtask. To evaluate consistency, we use the Mahalanobis distance between any latent state and the subtask distribution. For a latent state \( \mathbf{z}_t^{i,j} \in \mathbb{R}^d \) at timestep \(t\) in the \(j\)-th subtask of the \(i\)-th trajectory, the distance is computed as:

\begin{equation}
    d(\mathbf{z}_t^{i,j}, \mu^j, \Sigma^j) = \sqrt{\left( \mathbf{z}_t^{i,j} - \mu^j \right)^\top (\Sigma^j)^{-1} \left( \mathbf{z}_t^{i,j} - \mu^j \right)}
    \label{eqn:md}
\end{equation}

Here, \( \mu^j \) and \( \Sigma^j \) are the mean and covariance of the latent features for subtask \(j\), as defined earlier.

\textbf{Scoring Subtasks: }Given \(N \times k\) subtasks, we evaluate each by computing the Mahalanobis distance \(d\) at every timestep and averaging across the subtask. Higher scores indicate greater deviation from the nominal feature distribution. Subtasks are then ranked in descending order of their scores, and the top \(\rho\) most inconsistent segments are identified as corrupted. These segments can either be removed from the training set or assigned a weight of \(\beta_{ij} = 0\) in the loss function defined in Equation~\ref{eqn: prob2}.  This subtask-level scoring also helps correct misclassifications from earlier stages. For instance, a trajectory labeled as "good" may still contain a localized failure. Our method can isolate and penalize such segments appropriately. Notably, it is not the absolute magnitude of \(d\), but its temporal behavior that provides insight: an increasing trend suggests the robot is drifting out-of-distribution, while a decreasing trend indicates recovery toward the nominal feature distribution.


\section{Experimental Set-ups}

Through our experiments, we aim to demonstrate that even poor-quality demonstrations from non-expert users can contain valuable segments.  We assess how filtering out the low-quality segments, based on GiB's predictions, impacts downstream policy performance. To highlight the utility of our approach, we focus on multi-step long-horizon robotic manipulation tasks.

\begin{figure}[thpb]
  \includegraphics[width=0.45\textwidth, height=3.0cm]{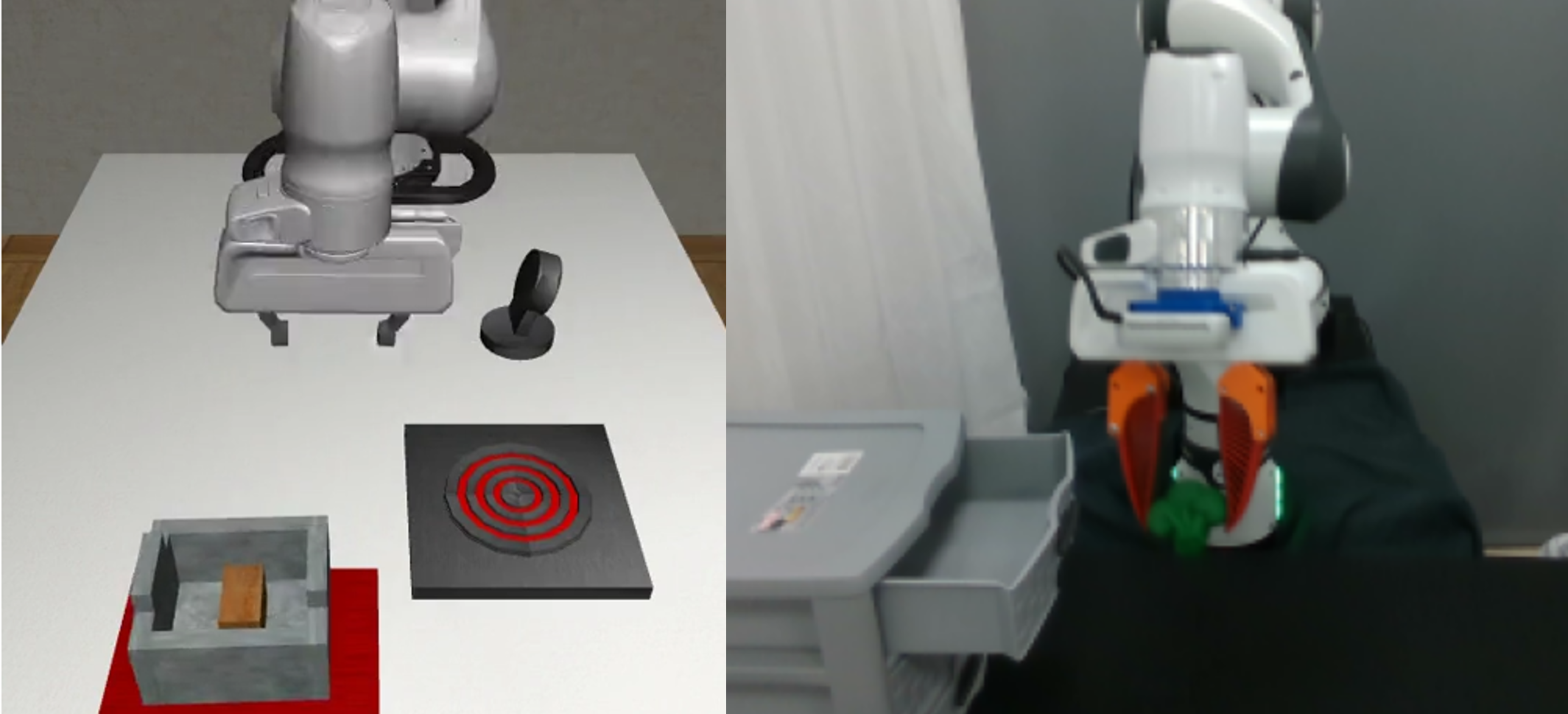}
  \caption{Simulation task (Kitchen, left) and real-world Franka task (right).}
  \label{fig:realtask}
\end{figure}

\subsection{Tasks:} For simulation, we selected tasks from the MimicGen environment~\cite{mandlekar2023mimicgen}, covering varying complexity and horizons. We used four tasks: \textbf{Square}, a two-step task where the robot picks up a nut and places it on a rectangular peg; \textbf{Coffee}, a two-step task involving inserting a coffee pod into a machine and closing the lid; \textbf{Mug Cleaning}, a three-step task requiring the robot to open a drawer, pick up a mug, place it inside, and close the drawer; and \textbf{Kitchen}, a long-horizon task composed of seven sequential pick-and-place subtasks. For real-world experiments, we designed a task similar to \textsl{Mug Cleaning}, where a Franka robot picks up an a broccoli from a tabletop (\textbf{Subtask 1}), places it into an open drawer (\textbf{Subtask 2}), and then closes the drawer (\textbf{Subtask 3}). Figure \ref{fig:realtask} shows an example of the kitchen task and the real robot drawer task.

\subsection{Demonstration Set:} Each demonstration captures state and action data. Simulation states include two camera images (\(\mathbb{R}^{84 \times 84 \times 3}\)), end-effector position (\(\mathbb{R}^3\)), end-effector rotation (\(\mathbb{R}^4\)), and gripper status (\(\mathbb{R}^2\)). Real-world states are similarly structured but with higher-resolution images (\(\mathbb{R}^{320 \times 240 \times 3}\)), end-effector states (\(\mathbb{R}^{16}\)), joint positions (\(\mathbb{R}^7\)), and gripper status (\(\mathbb{R}^1\)). In both cases, actions are represented by a \(\mathbb{R}^7\) vector, comprising six delta end-effector movements and one binary gripper command.

\textbf{Non-Expert Demonstration Collection:} Demonstrations were collected using a SpaceMouse interface in an IRB-approved study involving one expert (an author) and five non-expert participants (3 male, 2 female; ages 31–61, mean 44, SD 11), none with gaming or technical backgrounds. Participants were not instructed to make mistakes; errors arose naturally. After a brief tutorial, each provided 10–20 demonstrations based on comfort. Some demonstrations were incomplete, containing only a subset of subtasks.

\textbf{Dataset:} Collecting demonstrations from lay users presented typical FIL challenges, long-horizon tasks led to fatigue, and many demos were heavily corrupted, often yielding 0\% policy success. To systematically evaluate GiB's effectiveness, we constructed the following datasets. \(\mathbf{D_{12}}\): An unlabeled dataset combining expert and lay user demonstrations, used to compare policy performance after applying GiB. It includes 40 expert and 30 lay demos for Coffee and Square, 40 expert and 20 lay for Mug, and 30 expert and 20 lay for Kitchen.  
\(\mathbf{D_{22}}\): Real-world data with 40 high-quality and 20 low-quality demonstrations from two authors, the latter mimicking common user errors. Collecting real-world corrupted demos from lay users poses safety risks and is thus approximated through controlled collection.

\subsection{BED model and Baselines}
\textbf{BED Policy $\hat{\pi}_\theta$:} We use BED as a pretext task to learn latent representations and trajectory weights. The architecture follows behavioral cloning (BC), with a ResNet-18 vision encoder for each image input and an MLP for each low-dimensional input. These features are fused via a three-layer MLP, action decoder. We build on RoboMimic’s \cite{robomimic2021} BC implementation and introduce a learnable weight parameter \( w \) for each trajectory.


\textbf{Baselines:} We assess GiB's impact by applying it to mixed-quality data prior to training Diffusion Policy \cite{chi2024diffusionpolicy}, a state-of-the-art imitation learning method, in both simulation and real-robot settings. We use default hyperparameters, training Diffusion Policy for 200 epochs in simulation and 800 epochs on the real robot task. For the other baselines (S2I \cite{chen2024towards}, LOF, and DemInf\cite{hejna2025robot}), we generated masks and applied them in the same way as the GiB mask.


\section{Experiments and Results}
We evaluated the effectiveness of GiB in both simulation and real-world experiments by training policies on datasets filtered with GiB as well as on datasets filtered with alternative methods, and we compared their performance in terms of downstream task success. All models were trained with three different seeds, and the reported success rates correspond to the average across these runs.

\subsection{Simulation Results:}  
\begin{table*}[htbp]
  \centering
    \caption{Average success rate of Diffusion Policy obtained from 3 runs with different seeds and evaluated in simulation every 20 epochs.}

  \begin{tabular}
  {|p{1.5cm}|p{1.5cm}|p{1.5cm}|p{1.5cm}|p{1.5cm}|p{1.5cm}|p{1.5cm}|p{1.5cm}|}
    \hline
     & All data & Oracle masked & BED masked& DemInf masked& S2I &LOF masked & GiB masked \\
    \hline
    Square & $0.34 \pm 0.0$ & $0.45 \pm 0.04$ &  $0.45 \pm 0.04$  & $0.45 \pm 0.04$ & $0.23 \pm 0.02$ & $0.37 \pm 0.02$ &$\textbf{0.49} \pm 0.02$ \\
    \hline
    Mug & $0.73 \pm 0.09$ & $0.79 \pm 0.02$ & $0.79 \pm 0.02$ & $0.75 \pm 0.02$ & $0.55 \pm 0.03$ & $0.76 \pm 0.02$ &  $\textbf{0.79} \pm 0.04$ \\
    \hline
    Coffee & $0.73 \pm 0.02$ & $0.81 \pm 0.05$ & $0.74 \pm 0.02$ & $\textbf{0.77} \pm 0.02$ & $0.1 \pm 0.00$ & $\textbf{0.77} \pm 0.04$ & $0.76 \pm 0.03$\\
    \hline
    Kitchen & $0.79 \pm 0.05$ & $0.77 \pm 0.01$ & $0.87 \pm 0.15$ & $0.79 \pm 0.06$ & $0.69 \pm 0.04$ & $0.81 \pm 0.02$ & $\textbf{0.87} \pm 0.04$\\
\hline
\textbf{Mean} 
& $0.65 \pm 0.04$
& $0.71 \pm 0.03$
& $0.71 \pm 0.06$
& $0.69 \pm 0.04$
& $0.39 \pm 0.02$
& $0.68 \pm 0.03$
& $\textbf{0.73} \pm 0.03$ \\
\hline
  \end{tabular}
  \label{tab:policy_comparison}
\end{table*}

Table \ref{tab:policy_comparison} reports the performance of Diffusion Policy when trained on datasets filtered by GiB and five alternative methods. The results show that GiB consistently delivers competitive or superior performance across a diverse set of tasks. Notably, GiB achieved the highest success rate on the Square task (0.49) and matched the top performance on the Kitchen task (0.87). Its result on the Mug task (0.79) was equivalent to the top-performing baselines, Oracle- and BED-masked. In this case, BED filtering successfully identified the ground-truth bad demonstrations, resulting in the same training dataset for both variants. While GiB slightly trailed the Oracle-masked variant on the Coffee task (0.76 vs. 0.81), a key trend emerges: GiB’s strength is most pronounced in complex scenarios. Unlike Oracle’s aggressive ground-truth pruning, which can hinder learning on difficult tasks by discarding valuable data, GiB’s more conservative strategy of retaining high-quality segments from otherwise imperfect demonstrations provides a robust advantage, leading to stronger performance where data efficiency is critical.




\begin{table}[h]
\begin{center}
\centering
\caption{Average success rates of Diffusion Policy on the Franka Drawer task. For BED, GiB, DemInf-masked, and LOF-masked, results are averaged over 40 rollouts. For the remaining two methods, 20 rollouts were used given their consistent performance across trials.}
\label{table:realrobot_exp}
\begin{tabular}{|c|c|c|c|c|}
\hline
\textbf{Dataset}  & Sub1 & Sub2 & Sub3 & Full Task \\
\hline
Oracle & 0.45 & 0.85 & 0.55 & 0.15 \\
\hline
All data & 0.20 & 0.75 & 0.35 & 0.15 \\
\hline
BED-masked & 0.55 & 0.95 & 0.70 & 0.45 \\
\hline
DemInf-masked & 0.575 & 0.875 & 0.8	& 0.4 \\
\hline
LOF-masked & 0.625 & 0.7 & 0.675 & 0.275 \\
\hline
GiB-masked & \textbf{0.78} & \textbf{0.98} & \textbf{0.93} & \textbf{0.70} \\
\hline
\end{tabular}
\end{center}
\end{table}


\subsection{Real-World Results:}  

We evaluated GiB on the real-world dataset \( \mathbf{D_{22}} \) using Diffusion Policy. To assess GiB’s effectiveness in handling noisy demonstrations, we compared it against three masking-based baselines: BED, DemInf, and LOF. Table~\ref{table:realrobot_exp} reports success rates for each subtask and the full task. Training on only good demonstrations (Only Good) yielded strong subtask performance (e.g., 0.85 on Sub2) but poor full-task success (0.15). Including 20 corrupted demonstrations (All) reduced performance in some subtasks.  

All masking-based baselines outperformed both the Only Good and All datasets, showing higher subtask performance and improved full-task success. This highlights the importance of curating the dataset: poor-quality demonstrations can degrade behavior and lower success rates. However, the method of pruning also affects results. GiB-masking was able to outperform all other baselines, achieving the highest subtask scores and a full-task success of 0.70, demonstrating both the value of proper data selection and GiB’s robustness in handling noisy demonstrations.

In our real-robot experiments, the S2I-masked method yielded a zero success rate, so we excluded it from Table \ref{table:realrobot_exp}. S2I requires robot-specific data in addition to the imitation-learning data used by all other methods. In particular, it depends on precise camera calibration data, which we did not have; instead, we used the default calibration values provided with the RealSense camera. These defaults are not generally accurate, which we suspect is the reason behind the poor performance.


\subsection{Interpreting GiB:} 

\begin{figure}[thpb]
  \includegraphics[width=0.48\textwidth, height=3.0cm]{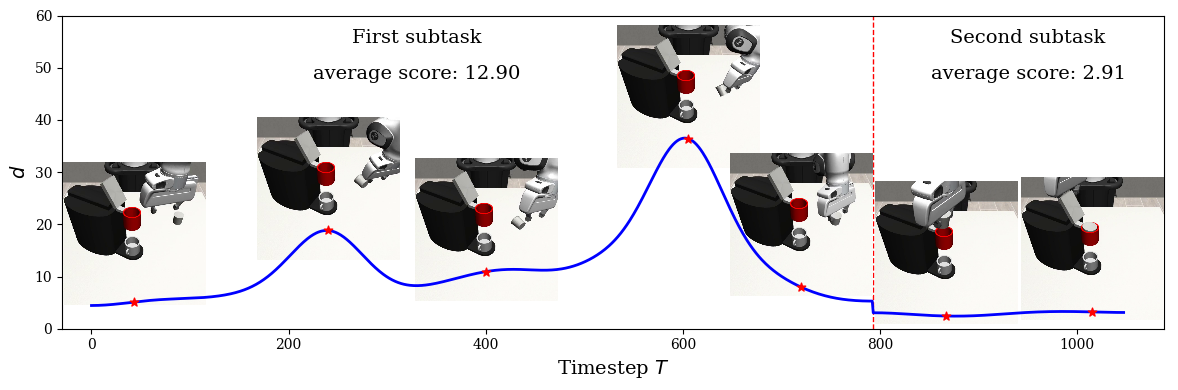}
  \caption{Visualization of a coffee-making task where the first subtask is executed incorrectly and the second correctly. The red dotted line separates the first and second subtask. The blue line shows the Mahalanobis distance at each timestep from the corresponding subtask’s feature distribution. Higher values indicate greater deviation from expected behavior.}
  \label{fig:goodbad}
\end{figure}

Figure~\ref{fig:goodbad} illustrates GiB's ability to assess subtask execution quality in a two-step coffee-making task: (1) picking a coffee pod and (2) inserting it into the machine. The plot shows the Mahalanobis distance of each timestep’s latent feature from the corresponding subtask distribution. Seven representative frames are shown. The first subtask exhibits consistently high distances, indicating deviation from nominal behavior. The second subtask shows lower distances, suggesting it was executed correctly. A peak at timestep 605 corresponds to a rare state in which the robot nudges the pod upright, a behavior that is not well represented in the dataset. While this moment shows high deviation, the decreasing trend over the next 200 timesteps indicates recovery. This example demonstrates GiB's interpretability: it not only flags deviations but also reveals when the robot successfully recovers, offering insight into demonstration quality over time.


\subsection{Multimodal Experiment:}  

\begin{wrapfigure}{l}{0.25\textwidth} 
    \vspace{-5pt} 
    \includegraphics[width=0.25\textwidth]{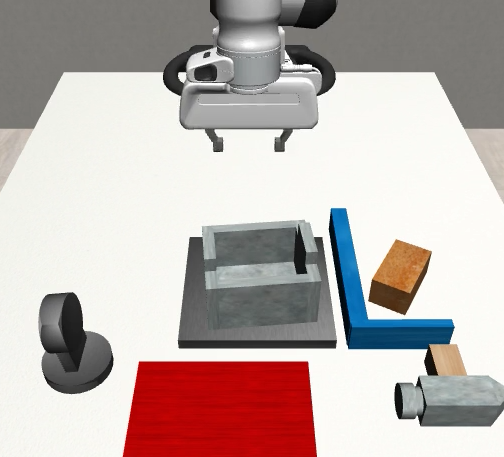}
    \caption{Multimodal kitchen environment.}
    \label{fig:multimodal}
    \vspace{-10pt} 
\end{wrapfigure}

\begin{figure}[thpb]
  \includegraphics[width=0.5\textwidth]{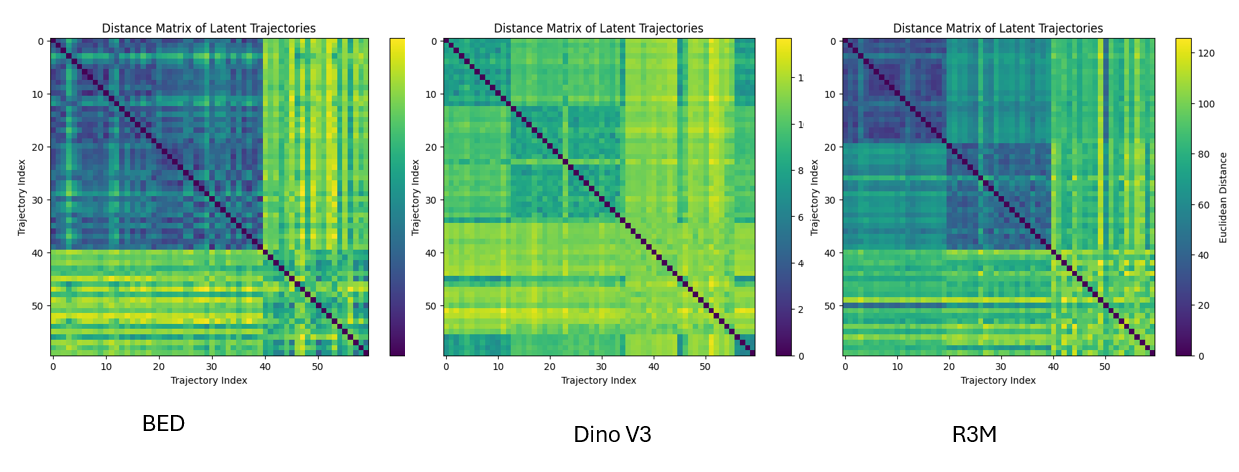}
  \caption{Even though the task can be performed in two distinct modes, BED learns a unified latent representation—grouping both modes together as the same task while clearly separating poor or erroneous demonstrations. In contrast, other representation methods split the two valid modes into separate clusters, in the same way they represent bad demonstrations as a separate group.}

  \label{fig:mode}
\end{figure}

To demonstrate multimodal way of performing the same task, we created another Kitchen environment by modifying the Kitchen environment in MimicGen. It is a two-step task with added additional features to the environment. The task requires picking up the pot from the stove and placing it on the serving region (the red square on the table). To increase diversity, we collected multimodal demonstrations, where the pick-up subtask involved grasping the object from different sides. We collected a balanced demonstration set of 60 demos where in 20 demos the robot picked the left handle of the pot, in another 20 demos the robot picked right handle of the pot and 20 erroneous demonstrations. To evaluate robustness under multimodal demonstrations, we applied both our method and baseline approaches. We observed that both BED and DemInf
successfully identified all suboptimal demonstrations. While the BED loss 
function is designed to learn a nominal goal and a nominal path, it might 
initially appear unsuitable for multimodal demonstrations. However, BED first 
learns a representation that explicitly accounts for multimodality. As a 
result, it classifies demonstrations as ``bad'' only when they exhibit 
inconsistency, whereas variations such as executing the task from the left 
or right remain consistent. Moreover, the multiple terms in the BED objective 
further stabilize learning, enabling it to handle multimodal 
demonstrations effectively. Figure \ref{fig:mode} shows that, although the task can be performed in two distinct modes, BED learns a unified latent representation that groups both modes together as the same task while clearly separating poor or erroneous demonstrations; in contrast, other representation methods split the valid modes into separate clusters, treating them similarly to how they isolate bad demonstrations into a distinct group.



\subsection{Tuning Hyperparameter $\rho$:}  
\begin{figure}[thpb]
  \includegraphics[width=0.5\textwidth]{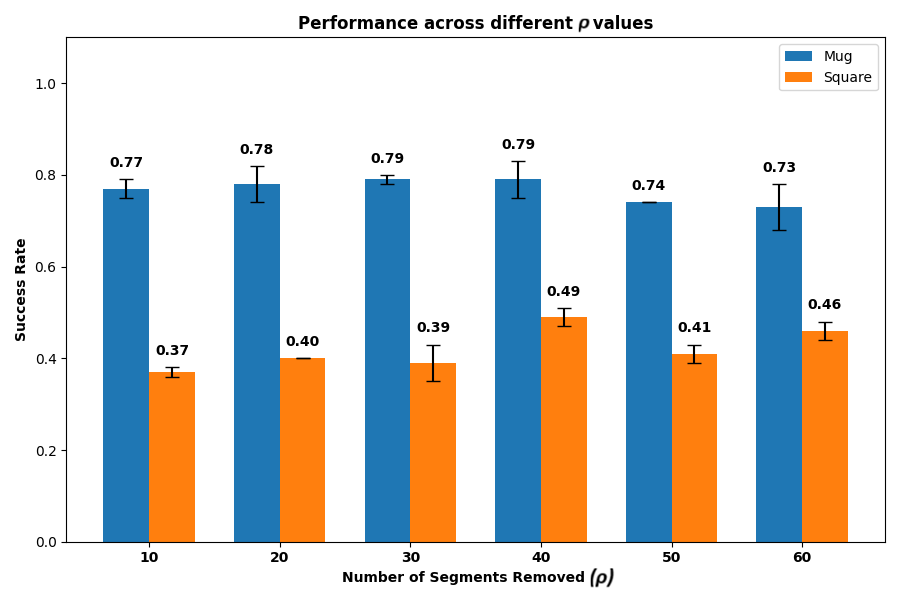}
  \caption{Success rate under different $\rho$ values for Square and Mug tasks.}

  \label{fig:hyper-r}
\end{figure}

Fig. \ref{fig:hyper-r} shows how the pruning parameter $\rho$ influences the success rate of the policy. In the square task, the best performance occurs at $\rho=40$. Pruning beyond this point removes too much useful data, shrinking the dataset and lowering success. Conversely, pruning below 40 retains poor demonstrations in the dataset, which also harms performance. The mug task, being simpler, shows the same overall trend but less strongly. Because it is easier to learn, removing data at 30 or 40 has little impact. However, including poor demonstrations or pruning too aggressively (at 50 and 60) reduces success rates.



\section{Conclusion and Future Work}

In this paper, we introduced \textit{GiB}, a method for automatically identifying high- and low-quality subtasks within non-expert demonstrations. By filtering out suboptimal trajectory segments, \textit{GiB} addresses a key challenge in Field Imitation Learning (FIL): learning robust policies from imperfect human data. Experiments with demonstrations collected from five non-expert users showed that \textit{GiB} consistently improves policy performance compared to unfiltered data and alternative filtering techniques. These results underscore the value of non-expert demonstrations \textit{if} paired with effective filtering like \textit{GiB}, which ensures data quality and enhances policy reliability.

Looking forward, while \textit{GiB} does not explicitly identify recovery behaviors, its subtask-level deviation scores naturally reveal them. For instance, in Fig.~\ref{fig:goodbad}, a spike in deviation highlights an error in subtask 1, followed by a recovery phase where the demonstrator corrects the mistake, enabling successful completion of subtask 2. We propose extending \textit{GiB} to automatically detect and utilize such recovery phases during training. Incorporating these corrective strategies would enable policies to learn not only successful actions but also how to recover from mistakes, an essential capability for real-world generalization.



\section{Limitations}



Our method assumes that each demonstration can be segmented into a fixed number of subtasks, which are generated using heuristic-based scripts. This segmentation strategy can break down when users repeat or revisit actions, such as retrying a failed step, leading to misaligned segments. To overcome this, we plan to explore learning-based segmentation approaches, such as those leveraging semantic or temporal structure, which could reduce reliance on task-specific heuristics and improve generalization across diverse behaviors.





\bibliographystyle{IEEEtran}
\bibliography{subopt.bib}

\end{document}